\documentclass[sigconf]{acmart}
\usepackage{natbib}
\usepackage{ulem} 
\usepackage{multirow}
\usepackage{pgfplots}
\usepackage{pgfplotstable}
\pgfplotsset{width=7.5cm,compat=1.13}
\AtBeginDocument{%
  \providecommand\BibTeX{{%
    \normalfont B\kern-0.5em{\scshape i\kern-0.25em b}\kern-0.8em\TeX}}}

\setcopyright{acmcopyright}
\copyrightyear{2020}
\acmYear{2020}
\acmDOI{10.1145/1122445.1122456}

\acmConference[KDD '20]{KDD '20}{August 22--27, 2020}{San Diego, CA, USA}
\acmBooktitle{KDD '20: The 26th ACM SIGKDD Conference on Knowledge Discovery and Data Mining, August 22--27, 2020, San Diego, CA, USA}
\acmPrice{15.00}
\acmISBN{978-1-4503-XXXX-X/18/06}

\newcommand{\vpara}[1]{\vspace{0.05in}\noindent\textbf{#1 }}
\newcommand{\interbert}{InterBERT}

\begin{document}

%%
%% The "title" command has an optional parameter,
%% allowing the author to define a "short title" to be used in page headers.
\title{InterBERT: Vision-and-Language Interaction for Multi-modal Pretraining}

%%
%% The "author" command and its associated commands are used to define
%% the authors and their affiliations.
%% Of note is the shared affiliation of the first two authors, and the
%% "authornote" and "authornotemark" commands
%% used to denote shared contribution to the research.
% \author{Junyang Lin}
% \authornote{Both authors contributed equally to this research.}
% \email{junyang.ljy@alibaba-inc.com}
% \orcid{1234-5678-9012}
% \author{An Yang}
% \authornotemark[1]
% \email{yangan.ya@alibaba-inc.com}
% \affiliation{%
%   \institution{Alibaba Group}
%   \streetaddress{P.O. Box 1212}  
%   \city{Dublin}
%   \state{Ohio}
%   \postcode{43017-6221}
% }

\author[J. Lin*, A. Yang*, Y. Zhang, J. Zhou, H. Yang]{
    Junyang Lin$^{*1}$, An Yang$^{*1,2}$, Yichang Zhang$^1$, Jie Liu$^1$, Jingren Zhou$^1$, Hongxia Yang$^{\dagger 1}$
}
\affiliation{
    $^1$ Alibaba Group
}
\affiliation{
  $^2$ MOE Key Lab of Computational Linguistics, School of EECS, Peking University
}
\email{
  {junyang.ljy, yichang.zyc, sanshuai.lj, jingren.zhou, yang.yhx}@alibaba-inc.com
}
\email{
  yangan@pku.edu.cn
}

\begin{abstract}
Multi-modal pretraining for learning high-level multi-modal representation is a further step towards deep learning and artificial intelligence. In this work, we propose a novel model, namely \interbert~(BERT for Interaction), which is the first model of our series of multimodal pretraining methods M6 (MultiModality-to-MultiModality Multitask Mega-transformer). The model owns strong capability of modeling interaction between the information flows of different modalities. The single-stream interaction module is capable of effectively processing information of multiple modalilties, and the two-stream module on top preserves the independence of each modality to avoid performance downgrade in single-modal tasks.  
We pretrain the model with three pretraining tasks, including masked segment modeling (MSM), masked region modeling (MRM) and image-text matching (ITM); and finetune the model on a series of vision-and-language downstream tasks. Experimental results demonstrate that \interbert~outperforms a series of strong baselines, including the most recent multi-modal pretraining methods, and the analysis shows that MSM and MRM are effective for pretraining and our method can achieve performances comparable to BERT in single-modal tasks. 
Besides, we propose a large-scale dataset\footnote{We will release the dataset to nourish further development in the community.} for multi-modal pretraining in Chinese, and we develop the Chinese \interbert~which is the first Chinese multi-modal pretrained model. We pretrain the Chinese \interbert~on our proposed dataset of 3.1M image-text pairs from the mobile Taobao, the largest Chinese e-commerce platform. We finetune the model for text-based image retrieval, and recently we deployed the model online for topic-based recommendation.
%  and the online performance demonstrates its industrial potential in e-commerce. 
\end{abstract}


% \ccsdesc[500]{Computer systems organization~Embedded systems}
% \ccsdesc[300]{Computer systems organization~Redundancy}
% \ccsdesc{Computer systems organization~Robotics}

%%
%% Keywords. The author(s) should pick words that accurately describe
%% the work being presented. Separate the keywords with commas.

%% A "teaser" image appears between the author and affiliation
%% information and the body of the document, and typically spans the
%% page.
% \begin{teaserfigure}
%   \includegraphics[width=\textwidth]{sampleteaser}
%   \caption{Seattle Mariners at Spring Training, 2010.}
%   \Description{Enjoying the baseball game from the third-base
%   seats. Ichiro Suzuki preparing to bat.}
%   \label{fig:teaser}
% \end{teaserfigure}

%%
%% This command processes the author and affiliation and title
%% information and builds the first part of the formatted document.
\maketitle

\renewcommand{\thefootnote}{\fnsymbol{footnote}}
\footnotetext[1]{Equal contribution. This work is done when An Yang is an intern at Alibaba Group.}
\footnotetext[2]{Corresponding author.}
\renewcommand{\thefootnote}{\arabic{footnote}}

\section{Introduction}
\label{sec:intro}

% In recent years, a focus in the community is the learning of multi-modal representation, especially vision-and-language representation learning. 
% The understanding of vision and language is a bedrock of cross-modal downstream tasks, such as visual commonsense reasoning~\cite{VCR}, image-text retrieval~\cite{image_text_retrieval}, etc. 
% However, in spite of the steady progress in the tasks, the methods are mostly based on supervised learning, and the scale of training data limits their performance. 
% It is hard to generalize the task-specific models to multiple visio-linguistic tasks. 
% Therefore, a number of researchers in this field have turned their focus to pretraining, which can generate generic representations.

Pretraining has raised much attention in the community due to its strong capability of generalization and efficient usage of large-scale data. The development of computer vision has been highly connected with pretraining, such as AlexNet~\cite{alexnet}, VGG~\cite{vgg} and ResNet~\cite{resnet}, which are pretrained on the large-scale dataset ImageNet~\cite{imagenet} for image classification. 
Recent years have witnessed the burst of pretraining in natural language processing. 
Pretrained models~\cite{elmo, ulmfit, bert, roberta, xlnet, unilm} have reached state-of-the-art performances in many downstream tasks of natural language processing (NLP), including question answering~\cite{squad}, natural language inference~\cite{glue}, and even natural language generation, such as neural machine translation~\cite{seq2seq, attention, transformer} and abstractive summarization~\cite{abs, ibmsum}. 

% the progress of multi-modal pretraining
Such significant progress in this field raised the concern of pretraining for task-agnostic multi-modal representation. 
A series of cross-modal pretraining methods were proposed, and the self-supervised learning provides the models with a strong ability to adapt to multiple multi-modal downstream tasks through finetuning~\cite{videobert,visualbert,lxmert,vilbert,vl-bert,uniter,unicoder-vl,vlp,mil-nce,oscar}. However, these models are mostly pretrained by simple tasks such as masked language/object modeling (MLM/MOM) and image-text matching (ITM). 
Except for that, single-stream models~\cite{vl-bert,uniter,unicoder-vl,oscar} simply apply BERT and mix information from two streams into one model, while two-stream models~\cite{vilbert,lxmert} can only build interaction with co-attention, where there is no self attention to the self-context in each layer of co-attention. 

Motivated by this observation, we propose a novel method for multi-modal pretraining, called \textbf{\interbert}, which refers to \textit{\textbf{BERT} for \textbf{Inter}action}. This model is the first one of our series of pretraining methods M6 (MultiModality-to-MultiModality Multitask Mega-transformer). 
The proposed architecture consists of a single-stream interaction module for all the inputs from different modalities, as well as a two-stream extraction module that processes information from each modality separately. 
This architecture ensures sufficient interaction between modalities and generates contextualized mode representations. 
Besides, we pretrain the model with our proposed masked group modeling (MGM) and image-text matching with hard negatives (ITM-hn). The tasks are more challenging as they force the model to predict a span or a region and differentiate positive and negative samples with higher difficulty, which requires the model to build a stronger connection between modalities. 

We pretrain our \interbert~on a series of large-scale datasets of image-text pairs, 
and we evaluate the effects of \interbert~on several multi-modal downstream tasks, including caption-based image retrieval~\citep{image_text_retrieval}, zero-shot caption-based image retrieval, and visual commonsense reasoning~\citep{VCR}. 
Experimental results demonstrate that our method can achieve significant improvements over the baseline models, and it outperforms or rivals the recent multi-modal pretrained models. 
% We also evaluate the effects of our pretraining tasks and the model performance in single-modal tasks. 
The analysis demonstrates that our pretraining tasks can positively impact the model performance in different downstream tasks, and the model can adapt to single-modal tasks without significant performance decrease in comparison with the BERT-base model. 
% We also find that the weight initialization for pretraining can make a difference in the finetuning of certain downstream tasks. 
Furthermore, we deploy the model in the mobile Taobao, the largest platform of e-commerce in China. We conduct an A/B test and achieve improvements in click-through rate and exposed category width over the single-modal baseline based on BERT. This demonstrates that the potential of pretraining and the contribution of cross-modal information in recommendation.

In brief, our contributions are illustrated below:
\begin{itemize}
    \item We propose a novel method for multi-modal pretraining called \textbf{\interbert}. The new architecture effectively builds multi-modal interaction and preserves the independence of single-modal representation. The proposed pretraining task MGM encourages the model to learn group prediction and interaction with larger context, and the proposed ITM-hn enhances the difficulty in distinguishing positive and negative samples so as to improve its capability in cross-modal matching. 
    % We pretrain the model with a new series of pretraining tasks, including MSM, MRM and ITM, to improve the model performance on the downstream tasks. 
    \item Experimental results demonstrate that \interbert~enhances the performance in the downstream tasks, including image retrieval and VCR, in comparison with the current pretraining methods. Our ablation studies have shown the effects of our architecture in the adaptation to NLP tasks and the significant effects of our proposed MGM and ITM-hn in VCR and image retrieval. 
    % \sout{Our analysis shows that the proposed training techniques are beneficial to the model performance. }
    % \item Our analysis shows that this model has strong robustness and still perform better than conventional models trained by supervised learning, and does not suffer from performance downgrade compared with the other multi-modal pretraining methods.
    \item We deploy \interbert~in the mobile Taobao and conduct an A/B test in a traffic-intensive scenario of recommendation. Our multimodal pretrained model gains performance increases in several metrics over the single-modal BERT-based baseline. 
    % \sout{We pretrain \interbert~on a large-scale dataset of Chinese image-text pairs extracted from the mobile Taobao.} 
    %  We have deployed the model online for topic-based recommendation, which demonstrates its superiority for e-commerce.
    % \sout{The online performance demonstrates the industrial potential of \interbert.} 
\end{itemize}

\vpara{Organization}
The rest of the paper is organized as follows:
% Section 2 reviews the related work, with a focus on pretraining. 
Section 2 provides an overview of the proposed approach, including model architecture as well as the pretraining methods.
% Moreover, it introduces the necessary notations and definitions and illustrates the background of BERT and Transformer. 
Section 3 provides the details of our experiments, including those of datasets, implementation, results and analysis. 
Section 4 introduces our Chinese \interbert~and our online deployment. 
% Section 5 describes the proposed dataset TaoMultimodal and the implementation of our Chinese \interbert. 
Section 5 reviews the related work, and the final section concludes the paper.

\section{Approach}
\label{sec:approach}

We detail our proposed approach \textbf{\interbert}~in this section. Before moving into the introduction to \interbert, we illustrate the background of pretraining in NLP and extend it to multi-modal pretraining. 

\subsection{Background}
\label{sec:background}
While following the pretraining principle in NLP, we first introduce the background of NLP pretraining and further introduce multi-modal pretraining. 

We first introduce the background of NLP pretraining. Given an input text, a word (a character or a subword) sequence with classification and separation tokens $w=\{[CLS], w_1,w_2,\cdots,w_n, [SEP]\}$ of length $n+2$, the model should learn to generate its high-level representations $h=\{h_{[CLS]}, h_1,h_2,\cdots,h_n, h_{[SEP]}\}$. 
The input can be a sequence of multiple sentences, where they are separated by ``$[SEP]$''. 
For the word embedding, except for the embedding layer, positional embedding and segment embedding are applied to denote the word positions and the segments they are from. 
For a BERT of $l$ layers, the model can produce $l$ sequences of representations $H = \{H^1, H^2, \cdots, H^l\}$. 
In most cases, we refer $h$ to $H^l$. 
For further finetuning, the pretrained model except for the topmost layer for logits is applied as the backbone of the model for a specific downstream task. 

Following this logic, such pretraining can be extended to learning multi-modal representations. In this work, we focus on the pretraining of vision and language. The dataset for multi-modal pretraining consists of paired image-text data, such as an image and its caption. To feed an image into BERT, a solution is to extract the object representations and bounding boxes with a detector, such as Faster-RCNN~\cite{faster-rcnn}, and form a sequence of $m$ object representations $o = \{ o_1, o_2, \cdots, o_m\}$ together with their positions. 
Similar to the pretraining in NLP, we also add a representation $o_{[CLS]}$, which is the mean pooling of the original $o$ in our implementation. 
The goal of the model is to learn the high-level representations of both image and text $h = \{h^i, h^t\}$, where $h^i = \{h^i_{[CLS]}, h^i_1, \cdots, h^i_m\}$ and $h^t = \{h^t_{[CLS]}, h^t_1, \cdots, h^t_n, h^t_{[SEP]}\}$. 
% For finetuning, we add layers based on the specific downstream tasks. 

\subsection{Model overview}

\begin{figure*}[tb]
    \centering
    \includegraphics[width=0.8\linewidth]{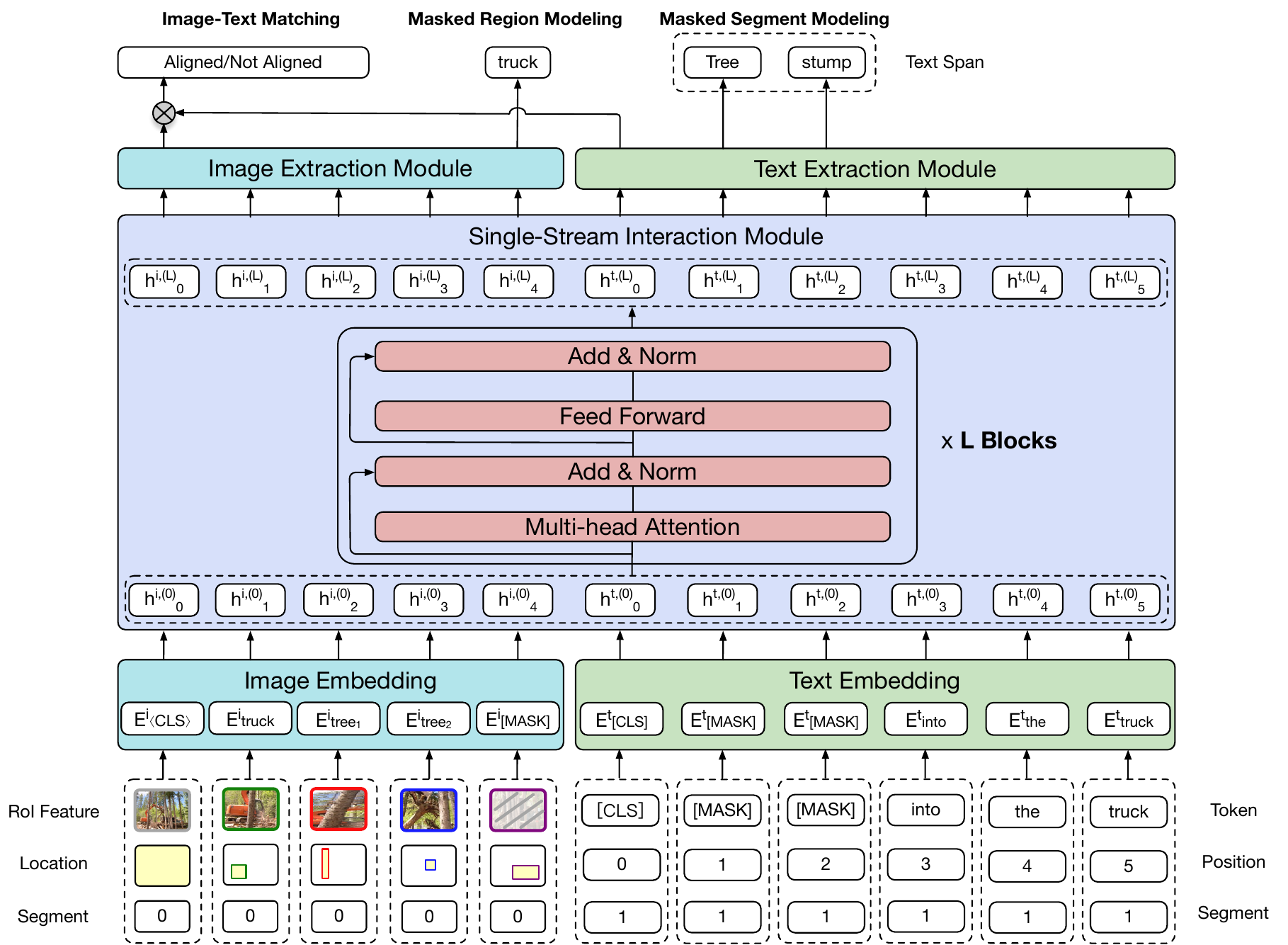}
    \caption{An overview of the architecture of \interbert. The model is built with an image embedding layer, a text embedding layer, a single-stream interaction module, and a two-stream extraction module.}
    \label{fig:model_overview}
  \end{figure*}

In this section, we illustrate the details of our proposed model \interbert. The overview of the architecture is demonstrated in Figure~\ref{fig:model_overview}. 
The simplest solution for multimodal pretraining is to pretrain a BERT-like model with the concatenation of image and text features. 
\citet{vilbert} pointed out that such a method of information fusing ignores the different requirements of processing for different modalities, and their experimental results show that the two-stream model outperforms the single-stream one in multiple tasks. 
We view that the effective interaction of modalities is the key to effective pretraining. Such interaction requires the gap bridging between image and text and the maintenance of the independence of each modality. 
Furthermore, an extra benefit of such independence enables transfer to both cross-modal downstream tasks and single-modal tasks. This enhances the robustness of the model and breaks the limitation of the form of pretraining data. 

\vpara{Replacing Co-Attention with All-Attention}
While co-attention demonstrates effects in VilBERT~\citep{vilbert}, we find that such a method of modal interaction limits the capability of the model. The representations of one modality can only attend to those of the other one, ignoring the self-context. The ideal attention should be one that attends to the whole context. Here we replace the co-attention with all-attention, which is a single-stream interaction module based on multi-head self attention (MHSA) and point-wise feed-forward neural network (FFN)~\citep{transformer,bert}. The input of the single-stream interaction module is the concatenation of image and text embeddings, and thus the attention can attend to the whole context of both modalities. 
The layer includes:
\begin{align}
    h^l &= \mathtt{MHSA}(\mathbf{W_{q}}x^{l-1}, \mathbf{W_{k}}x^{l-1}, \mathbf{W_{v}}x^{l-1}), \\
    \tilde{h^l} &= \mathtt{LN}(x^{l-1} + h^l), \\
    \hat{h^l} &= \mathbf{W_2}[\mathtt{GeLU}(\mathbf{W_1}\tilde{h^l} + b_1)] + b_2, \\
    x^l &= \mathtt{LN}(\tilde{h^l} + \hat{h^l}),
\end{align}
where $x_{l-1}$ is the whole context of image and text representations, instead of representations of a single modality. 
For multi-head attention, the model first transforms the inputs to query, key and value representations with weight matrices $\mathbf{W_q}$, $\mathbf{W_k}$, and $\mathbf{W_v}$, and split them into multiple heads and compute the attention scores of query and key as well as the weighted sum of the value. Layer normalization (LN) and residual connection are applied, and the activation function is GeLU~\citep{gelu}. 

This architecture enables strong interaction between modalities with the attention mechanism. 
Compared with the two-stream co-attention layer~\cite{vilbert} which can only attend to the representations of the other modality, this architecture enables a combination of self attention and co-attention, and therefore the model can generate more contextualized representations. Furthermore, another advantage is that the architecture is identical to BERT and thus its weights can be initialized with the pretrained BERT's weights, which improves the availability of the previous pretrained models. 

\vpara{Extraction Module for Mode Representations}
An ideal situation is that the model's outputs consist of visual and linguistic representations as well as the visual-linguistic ones. 
Also, a robust multi-modal pretrained model should have the capability to transfer to single-modal tasks. 
As mentioned above, the single-stream interaction module fuses the visual and linguistic representations and make them more contextualized. 
In concern of the extraction of representations of each modality, we should develop a module to respectively generate representations to separate the fused information. 

We implement a two-stream extraction module, which consists of an image extractor and a text extractor. Each extractor is based on self attention and FFN. 
The module is responsible for generating high-level object representations and text representations. 
Except for these, the model generates a general image representation and text representation for finetuning.
The image and text representations are transformed into a cross-modal representation by a multi-layer feed-forward network. 
To validate our hypothesis, we analyze by finetuning our pretrained model and the single-stream multi-modal pretrained model (a simple BERT architecture) on natural language processing tasks to evaluate their performances on single-modal tasks. 
The analysis demonstrates that our architecture can achieve similar performance compared with the original BERT-base model, while the single-stream model without the two-stream extraction module performs much worse. 
This shows our model's advantage in preserving modal independence. 
More details are described in Section~\ref{sec:analysis}.
% \yang{Can also be generalized to Multi-Stream Independence Module, right? If so, add one sentence for its further generalization.}

\vpara{Text Embedding and Image Embedding}
Following~\citet{bert}, we tokenize the input text and embed each word with an embedding layer. 
Positional embedding is required for the self-attention-based model to obtain the positional information, and segment embedding is required for the model to distinguish image and text. 

A solution to adapt the image to Transformer is to obtain the object representations and their locations with a detector. 
Following~\citet{vilbert}, we apply a commonly used object detector Faster-RCNN~\cite{faster-rcnn} trained on Visual Genome~\citep{visual_genome,bottom-up}. 
We extract the bounding boxes and the RoI (Region of Interest) features as the object representations. 
Similar to the aforementioned process, we apply positional embedding and segment embedding to the extracted features.

\subsection{Pretraining tasks}
In this section, we introduce the pretraining tasks for our multi-modal pretraining, namely masked group modeling (MGM) and image-text matching with hard negatives (ITM-hn). 

\vpara{Masked Group Modeling} 
We propose MGM, which encourages the model to predict the masked groups of images and texts. 
We name the masked group modeling on text ``masked segment modeling (MSM)'', and the masked group modeling on image ``masked region modeling (MRM)''. 
Similar to MLM, MSM also replaces the selected words with the same strategy (replacing with the token ``$\mathtt{[MASK]}$'', a random word or the original word). However, MSM masks a continuous segment of text instead of random words. Different from~\citet{spanbert}, we mask multiple segments for each sample. As to MRM, it masks selected objects with zero vectors as MOM does. Yet, it endeavors to mask objects that are immediate to avoid information leakage due to the overlapping between objects. MRM masks objects which have a high proportion of mutual intersection. 

For MSM, we randomly choose words as masking anchors by the probability of 10\%, and we randomly mask the anchors and 0 to 2 words after the anchors by the probability of uniform distribution. For MRM, we also randomly choose objects as masking anchors by the probability of 10\%, and we mask the objects whose IoUs with the anchors are larger than $0.4$ (the optimum value empirically). The objective of the model is to predict the masked words and the categories of the masked objects. The training minimizes the loss:
\begin{align}
    \mathcal{L}_{\mathbf{MSM}} &= -\mathbb{E}_{x \sim D} \log p\left(\overline{x}|\hat{x}\right) \approx -\frac{1}{N} \sum_{n=1}^{N} \sum_{t=1}^{T} \mathbf{m^{t}}(x^{n}, t) \log p_{\theta}\left(x_{t}^{n} | \hat{x}^{n}\right), \\
    \mathcal{L}_{\mathbf{MRM}} &= -\mathbb{E}_{x \sim D} \log p\left(\overline{x}|\hat{x}\right) \approx -\frac{1}{N} \sum_{n=1}^{N} \sum_{t=1}^{T} \mathbf{m^{i}}(x^{n}, t) \log p_{\theta}\left(x_{t}^{n} | \hat{x}^{n}\right),
\end{align}
where $x$ is a random sample of image-text pair from the training set $D$, and $\overline{x}$ refers to the masked segment or the masked region, and $\hat{x}$ refers the whole masked sequence $x$. $\mathbf{m^i}$ and $\mathbf{m^t}$ refer to the masking functions for image and text. The objective functions encourage the model to predict the masked groups of words or the class of the masked groups of objects.

\vpara{Image-Text Matching with Hard Negatives} 
For learning the relation between image and text, we regard the image-text pairs in the dataset as positive samples, and we pair the images with uncorrelated texts and regard the pairs as negative samples. Both the positive and negative samples share the same proportion in the training set. The model is pretrained to distinguish the positives and negatives. 

\begin{figure}
    \centering
    \includegraphics[width=0.8\linewidth]{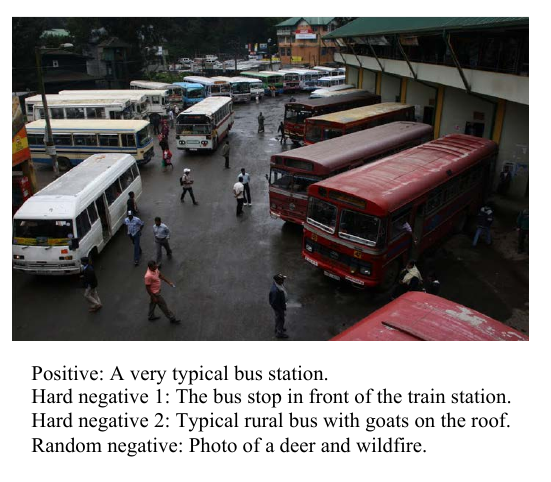}
    \caption{An illustration of the positive and negative samples of image-text matching. Based on TF-IDF similarity, we retrieve hard negative captions to increase the difficulty of pretraining.}
    \label{fig:example_itmhn}
\end{figure}

In previous works \cite{uniter, unicoder-vl, vilbert}, the uncorrelated captions in the negative samples are randomly selected from the training dataset, which makes the negatives easy to be distinguished out. To force the model to learn stronger cross-modal matching capability, we provide the model with harder negatives. For each training image, we consider the captions whose TF-IDF similarities with the image's original caption are lower than 0.5. The top 30 among them with the highest TF-IDF similarities are selected as the hard negative captions for the image. These operations are expected to select the captions which have more lexical overlaps with the positive caption but are semantically different. An example of hard negatives is shown in Figure~\ref{fig:example_itmhn}. During pretraining, we make 20\% negative samples constructed with the hard negative captions mentioned above and the other 80\% negatives still use randomly selected captions.\footnote{We have also attempted to set higher probability for the hard negative samples. However, this would make the pretraining loss to high for the model to converge.}

We add a simple MLP on top of the main architecture for computing the matching score between inputs of two modalities. Specifically, we first element-wisely multiply the image and text representations (the output representations at the position of ``$[CLS]$'') and send the generated representation through the MLP for the matching score. The training minimizes the cross-entropy loss:
\begin{align}
    \mathcal{L}_{\mathbf{ITM}} = -\mathbb{E}_{x,y \sim D}\left[y \log p\left(y|\hat{x}\right) + (1-y) \log \left(1-p\left(y|\hat{x}\right)\right)\right],
    % \max _{\theta} \quad \mathbb{E}_{\mathbf{z} \sim z_{T}}\left[\sum_{t=1}^{T} \log p_{\theta}\left(x_{z_{i}} | \mathbf{x}_{\mathbf{z}_{<t}}\right)\right]
\end{align}
where $x$ is a random sample from the training set $D$ and $y \in \{0,1\}$ denotes whether $x$ is positive or negative. $\hat{x}$ refers to the masked $x$. 

The overall objective function is the weighted sum of the aforementioned terms, as shown below:
\begin{align}
    \mathcal{L} &= \lambda _{1} \mathcal{L}_{\mathbf{MSM}} +  \lambda_{2} \mathcal{L}_{\mathbf{MRM}} + \lambda_{3} \mathcal{L}_{\mathbf{ITM}},
\end{align}
where $\lambda$ refers to the hyperparameter for the weights for each term.

\subsection{Finetuning}
We use the pretrained \interbert~as the backbone for the downstream tasks. We apply the pretrained model to three downstream tasks, including caption-based image retrieval, zero-shot caption-based image retrieval, and visual commonsense reasoning. Finetuning is simple as we can add simple MLP layers based on the requirements of the corresponding downstream tasks.
% \footnote{We demonstrate more implementation details about finetuning in Appendix~\ref{sec:appendix_impl}.}

\section{Experiments}
\label{sec:experiments}

In this section, we provide an introduction to our experimental details, and we demonstrate the results as well as the analysis.\footnote{The data statistics are presented in Table~\ref{tab:pretrain_dataset} and \ref{tab:downstream_datasets}.}
% More implementation details are referred to Appendix~\ref{sec:appendix_}.

% \subsection{Pretraining datasets}
% % % kddcup
% % Since the images of our downstream tasks involve different domains: photographic images (VCR and Flickr30K) and commodity pictures (KDD Cup), we separately prepare the pretraining data for these two types of tasks.
% For Flickr30K and VCR, we pretrain our model on the combination of three datasets, including Conceptual Caption (CC)~\citep{cc}, SBU Captions~\citep{sbu}, and COCO captions~\citep{coco}.\footnote{More details of pretraining data are in Appendix~\ref{sec:appendix_data}.}
% kddcup
% For the KDD Cup retrieval task, we have collected a pretraining dataset containing around 200 million pairs of product images and titles from the mobile Taobao \footnote{\url{https://www.taobao.com/}}, one of the world's largest e-commerce websites.
% We have removed the images that exist in the datasets of downstream tasks to avoid information leakage. 
% As shown in Table~\ref{tab:pretrain_dataset}, CC contains 3.3M image-caption pairs extracted from Google for training and 14K for validation. SBU contains 990K image-caption pairs for training and 10K for validation. 
% The length of each sentence is shorter than 36 words, and the length of each object sequence is shorter than 36 as we extract only 10 to 36 objects from each image following the previous work~\cite{vilbert}. 

\subsection{Downstream tasks}

\begin{table}[tb]
  \caption{Data statistics of the datasets for pretraining. The numbers in the parentheses refer to the numbers of images.}
  \label{tab:pretrain_dataset}
  \centering
  \begin{tabular}{ccc}
    \toprule
        Datasets&Training&Validation\\
        \midrule
        Conceptual Caption & 3.3M & 14K\\
        SBU & 890K & 10K \\
        COCO & 587K (117K) & 15K (3K) \\
    \bottomrule
  \end{tabular}
\end{table}

\begin{table}[tb]
  \caption{Data statistics of the datasets of the downstream tasks. ``i'' refers to the number of images, and ``t'' refers to the number of texts.}
  \label{tab:downstream_datasets}
  \centering
  \begin{tabular}{cccc}
    \toprule
        Datasets & Training & Validation & Testing\\
        \midrule
        Flickr30K & i:29K, t:145K & i:1K, t:5K & i:1K, t:5K \\
        % VQA & i:83K, t:444K & i:41K, t:214K & i:81K, t:448K \\
        VCR & i:80K, t:213K & i:10K, t:27K & i:10K, t:25K \\
        % RefCOCO+ & i:20K, t:141K & - & - \\
        % KDD Cup & i:3M, t:3M & i:15K, t:0.5K & i:30K, t:1K \\
    \bottomrule
  \end{tabular}
\end{table}

% To evaluate the effectiveness of \interbert, we conduct experiments on different downstream tasks, of which the images are under various domains. The Flickr30K and VCR datasets contain photographs in daily-life or movies, while the KDD Cup dataset is annotated on commodity pictures. More details of these tasks are provided below.

\vpara{Caption-Based Image Retrieval}
Caption-based image retrieval requires the model to retrieve an image from a large pool of images based on a given caption. We conduct experiments on Flickr30K~\citep{flickr}, whose images are extracted from Flickr.\footnote{\url{https://www.flickr.com}} In Flickr30K, each image is paired with five captions, which are of relatively high quality. 
Following~\citet{vilbert}, in the stage of training, we change the task to 4-way multiple choice by adding three negative images for each image-caption pair. 
The training set contains 29K images, and the validation and test set contain 1K images respectively. 
The evaluation metrics are R@1, R@5, and R@10 (recall at $1$, $5$, and $10$). 

\vpara{Zero-Shot Caption-Based Image Retrieval}
This is the zero-shot setting for caption-based image retrieval. The model performs caption-based image retrieval without finetuning on the training data. This challenges the capability of the pretrained model to understand the relations between image and text. 
% Zero-shot learning challenges the transferring ability to downstream tasks of the pretrained model. 
We use the same splits of the dataset and the same evaluation metrics as those of caption-based image retrieval. 

% \vpara{Visual Question Answering}
% Visual question answering is a task of cross-modal question answering~\cite{VQA}. 
% Given an image (or multiple images) and a question, the model should provide an answer. 
% This task requires the model to learn both visual information and linguistic information, which is more complex than conventional question answering. 
% In this work, we follow the previous work in multi-modal pretraining and implement our experiments on the dataset VQA 2.0~\cite{vqa2}. In this dataset, every question is related to a pair of images, which requires the model to pay sufficient attention to the visual information. The dataset is split into three subsets for training, validation and testing. The training set contains 83K images and 444K questions, the validation set contains 41K images and 214 questions, and the test set contains 81K images and 448K questions. The model should answer the questions by selecting answers from a shared set of 3129 answers. The evaluation metric is the VQA-score~\cite{vqa2}.

\vpara{Visual Commonsense Reasoning}
Visual commonsense reasoning (VCR) is a task connected with cognition and requires visual understanding~\cite{VCR}. There are three sub-tasks in VCR, including Q$\rightarrow$A, QA$\rightarrow$R, and Q$\rightarrow$AR. Q$\rightarrow$A refers to providing the answer based on the given image and question, QA$\rightarrow$R refers to providing the rationale based on the given image, question, and answer.
%Q$\rightarrow$AR refers to providing the answer and rationale based on the given image and question. 
% In QA$\rightarrow$R, the model should provide a rationale for the image, question, and answer. 
In Q$\rightarrow$AR, provided an image and a question, the model should not only answer the question but also give the correct rationale for the choice. 
For each question, there are 4 candidate answers and 4 candidate rationales. 
The training set contains 80K images and 213K questions, the validation set contains 10K images and 27K questions, and the test set contains 10K images and 25K questions. 
We apply accuracy score as the evaluation metric.

\subsection{Baselines}
For the comparison with the previous methods, we mainly compare our \interbert~with the previous models that achieved outstanding performances on the downstream tasks as well as the recent multi-modal pretrained models. 

\vpara{Previous Methods} For image retrieval, we compare \interbert~with SCAN~\cite{stacked_cross_attention}, which is an architecture based on stacked cross-attention. 
% For VQA, we compare \interbert~with the previous model, Bottom-Up Top-Down Attention~\cite{butd}. This is a model based on attention mechanism. 
For VCR, we compare \interbert~with R2C (Recognition to Cognition)~\cite{VCR}, which contains modules for grounding, contextualizing, and reasoning. 

\vpara{Multi-Modal Pretrained Models} We compare our model with some recent multi-modal pretrained models. Specifically, we focus on the comparison of our model with VilBERT and VL-BERT for the reason that our implementation details are similar to theirs, including pretraining datasets and the number of object features. 
Moreover, VilBERT and VL-BERT are regarded as powerful baselines of two-stream and single-stream multi-modal pretrained models, respectively\footnote{We focus on the comparison with the methods that released the codes and models for pretraining and finetuning. Both of them have released the full codes and the reported results proved reproducible. Please refer to \url{https://github.com/jiasenlu/vilbert_beta} and \url{https://github.com/jackroos/VL-BERT}.}. 
% \footnote{We follow the implementations of VilBERT that we use no more than 36 object features for each image, while Unicover-VL uses 100 features. 
% Furthermore, we also implemented a single-stream model with a similar architecture to UNITER and Unicoder-VL, which achieved results similar to those reported in VilBERT~\cite{vilbert}, which are different from the results of UNITER and Unicoder-VL. We leave it an open question for future discussion.}
% \footnote{Until recently, the codes of LXMERT, VilBERT and VL-BERT are released.}

\subsection{Implementation details}

% In the following, we introduce the details of our implementation of pretraining and finetuning, including the model architecture, optimizer, hyperparameters, etc. 

% \vpara{Pretraining}
We pretrain our model on Conceptual Caption (CC)~\citep{cc}, SBU Captions~\citep{sbu}, and COCO captions~\citep{coco}.
% \footnote{More details of pretraining data are in Appendix~\ref{sec:appendix_data}.}
For pretraining, we first extract the object representations of the images with a trained object detector. 
Specifically, the object representations and their bounding boxes are generated by an object detector based on Faster R-CNN~\cite{faster-rcnn} with a backbone of ResNet-101~\cite{resnet}, which is trained on Visual Genome~\cite{visual_genome}.\footnote{https://github.com/peteanderson80/bottom-up-attention}  
We pretrain the model with AdamW~\citep{adamw} with an initial learning rate of $1e-4$, $\beta_1=0.9$, $\beta_2=0.9999$, $e=1 \times 10 ^ {-6}$ and a weight decay of $0.01$. 
% We apply the linear decay learning rate scheduler with a warm-up period of $10000$ steps. 
% For more information, please refer to the appendix.
% \vpara{Finetuning}
For the finetuning on Flickr30K image retrieval, the maximum number of objects is $100$ and the actual numbers are between $90$ and $100$. 
% We use an initial learning rate of $4 \times 10 ^ {-5}$ and a batch size of $32$. 
The model reuses the output layer of the pretraining ITM task to compute the matching scores. 
We finetune the model on 8 Nvidia V100 for $20$ epochs with AdamW optimizer
with an initial learning rate of $4 \times 10 ^ {-5}$ and apply a linear decay learning rate scheduler with a warm-up period of $10000$ steps. 
For the finetuning on VCR, we use a smaller learning rate $2 \times 10 ^ {-5}$ and train the model for only $5$ epochs.
% \footnote{For more information about the implementation details, please refer to Appendix~\ref{sec:appendix_impl}.}
% For the finetuning on the KDD Cup query-based image retrieval, we finetune the model with a much smaller learning rate $5 \times 10 ^ {-6}$ for $20$ epochs, based on our preliminary experiments. During training, we construct $2$ negative samples for each ground-truth pair by randomly substituting the query or the image.

\subsection{Results}

\begin{table*}
  \caption{Results of the models on the three downstream tasks. The results of the baselines are those reported in their original papers. ``-'' denotes that the model was not implemented on the task in the original work. ``w/o pt'' refers to ``without pretraining''.}
  \label{tab:results}
  \begin{tabular}{c|ccccccccc}
    \toprule
    \multirow{2}{*}{Models} & \multicolumn{3}{c}{IR} & \multicolumn{3}{c}{Zero-shot}  & \multicolumn{3}{c}{VCR}\\
        & R@1 & R@5 & R@10 & R@1 & R@5 & R@10  & Q$\rightarrow$A & QA$\rightarrow$R & Q$\rightarrow$AR \\
        %  & & & val & test & val & test & val & test & & & & & & \\
        \midrule
        SCAN~\cite{stacked_cross_attention} & 48.6 & 77.7 & 85.2 & - & - & -   & - & -  & -  \\
        % BUTD & - & - & - & - & - & - & 65.3 & 65.7 & - & - & -  \\
        R2C~\cite{VCR} & - & - & - & - & - & - & 63.8 & 67.2 & 43.1 \\
        % MAttNet & - & - & - & - & - & - & - & - & - & - & - & - & - & - & 71.0 & 75.1 & 66.2\\
        \midrule
        % B2T2 & - & - & - & 55.0 & - & 72.6 & - & 75.7 & - & - & - & - & - & - \\
        VisualBERT~\cite{visualbert} & - & - & - & - & - & -  & 70.8 & 73.2 & 52.2 \\
        % LXMERT & - & - & - & - & - & - & 72.4 & 72.5 & - & - & - \\
        VilBERT~\cite{vilbert} & 58.2 & 84.9 & 91.5 & 31.9 & 61.1 & 72.8 & 72.4 & 74.5 & 54.0 \\
        VL-BERT~\cite{vl-bert} & - & - & - & - & - & - & \textbf{73.8} & 74.4 & 54.2 \\
        % Unicoder-VL & - & - & 54.5 & 54.9 & 72.6 & 73.4 & 74.5 & 74.4 & 71.5 & 90.9 & 94.9 & 48.4 & 76.0 & 85.2 & - & - & -\\
        % UNITER & 72.3 & 72.5 & - & 58.2 & - & 75.0 & - & 77.2 & 71.5 & 91.2 & 95.2 & 62.3 & 85.6 & 91.5 & 74.7 & 80.6 & 65.2\\
        \midrule
        \interbert~(w/o pt) & 53.1 & 80.6 & 87.9 & - & - & - & 63.6 & 63.1 & 40.3 \\
        \interbert & \textbf{61.9} & \textbf{87.1} & \textbf{92.7} & \textbf{49.2} & \textbf{77.6} & \textbf{86.0} & 73.1 & \textbf{74.8} & \textbf{54.9} \\
    \bottomrule
  \end{tabular}
\end{table*}

% We compare our model with both the previous models that reached SOTA before the application of multi-modal pretraining and the recent multi-modal pretrained models. In the following, we report the results and provide analyses. 

Table~\ref{tab:results} demonstrates the experimental results of our proposed model \interbert~as well as the compared baselines on the downstream tasks. 
In the experiment of image retrieval, \interbert~outperforms SCAN by a large margin (+13.3 (27.4\%) in R@1, +9.4 (12.1\%) in R@5, and +7.5 (8.8\%) in R@10), and it also outperforms VilBERT by +3.7 (6.4\%) in R@1, +2.2 (2.6\%) in R@5, and +1.2 (1.3\%) in R@10. As for zero-shot image retrieval, the advantage is significantly larger. It outperforms VilBERT by +17.3 (54.2\%) in R@1, +16.5 (27.0\%) in R@5, and +13.2 (18.1\%) in R@10). 
% In the experiment of VQA, \interbert~achieves an advantage of +5.3 (8.1\%) over BUTD and the performance is comparable to that of VilBERT. While LXMERT achieves a significantly better performance, we assume that this is because it has been pretrained on QA datasets, including Visual Genome and GQA. 
In the experiment of VCR, \interbert~also significantly outperforms the baseline R2C by +9.3 (14.6\%) in Q$\rightarrow$A, +7.6 (11.3\%) in QA$\rightarrow$R, and +11.8 (27.4\%) in Q$\rightarrow$AR, and it also outperforms VilBERT by +0.7 (1.0\%) in Q$\rightarrow$A, +0.3 (0.4\%) in QA$\rightarrow$R, and +0.9 (1.7\%) in Q$\rightarrow$AR. Compared with VL-BERT, \interbert~also outperforms by +0.7 (1.3\%) on the overall Q$\rightarrow$AR accuracy.

\interbert~has advantages over the baselines in the tasks, especially in zero-shot image retrieval. Also, compared with VilBERT, \interbert~has an advantage in the number of parameters (173M vs 221M), which reflects the effects of single-stream interaction. 
The significant advantage in zero-shot learning demonstrates that our model has a strong capability of modeling image-text relations and transferring to downstream tasks without finetuning. 

Also, we directly train our \interbert~without multi-modal pretraining to evaluate the effects of pretraining for the downstream tasks. To be more specific, as our pretrained model is initialized with the weights of BERT-base, we also train the \interbert~without pretraining with the BERT initialization. From Table~\ref{tab:results}, the model without pretraining suffers from the performance degrade (IR: -8.8 (-14.2\%) in R@1, -6.5 (-7.5\%) in R@5, and -4.8 (-5.2\%) in R@10; VCR: -9.5 (-13.0\%) in Q$\rightarrow$A, -11.7 (-15.6\%) in QA$\rightarrow$R, and -14.6 (-26.6\%) in Q$\rightarrow$AR). 
The effective multi-modal pretraining can significantly impact the model performance in downstream tasks. 

\subsection{Analysis}
\label{sec:analysis}

In this section, we conduct a series of analyses to evaluate the effects of MGM, the performance on single-modal downstream tasks, and the effects of weight initialization for pretraining.

\begin{table}[tb]
  \caption{An ablation study of the MGM conducted on the validation set of VCR.}
  \label{tab:ablation}
  \centering
  \begin{tabular}{cccc}
    \toprule
    \multirow{2}{*}{Tasks} & \multicolumn{3}{c}{VCR} \\
        & Q$\rightarrow$A & QA$\rightarrow$R & Q$\rightarrow$AR \\
        \midrule
       w/o MGM & 72.3 & 74.3 & 54.0  \\
       InterBERT & 73.1 & 74.8 & 54.9 \\
        % Two-stage & 70.3  \\
        % Out+In & MSM+MRM & - \\
        % Out+In+VCR & MSM+MRM & - \\
    \bottomrule
  \end{tabular}
\end{table}

\vpara{The Effects of MGM} We conduct an ablation study on the validation set of VCR to evaluate the effects of MGM, which includes MSM and MRM. Specifically, we pretrain two models with different pretraining tasks, including MLM+MOM+ITM and MSM+MRM+ITM. 
Table~\ref{tab:ablation} demonstrates the results of the evaluation. It can be found that our proposed MSM and MRM are beneficial to the pretraining effects. The model trained with MSM and MRM can outperform the baseline by $+0.8$ in Q$\rightarrow$A, $+0.5$ in QA$\rightarrow$R, and $+0.9$ in Q$\rightarrow$AR. The model trained with our tasks gains a stronger ability of modeling image and text by understanding contexts and building a stronger connection so that it can reach better performance in a task that requires reasoning over image and text. 
% However, directly pretraining with MSM and MRM damages the performance in the downstream task. We assume that the two tasks are relatively difficult for a model that is randomly initialized to learn, and therefore a hot start with a model first pretrained with relatively easy MLM and MOM can enhance the performance. 

\begin{table}[tb]
  \caption{An ablation study of our image-text matching conducted on the test sets of caption-based image retrieval and its zero-shot version.}
  \label{tab:ablation-itm}
  \centering
  \begin{tabular}{ccccccc}
    \toprule
    \multirow{2}{*}{Tasks} & \multicolumn{3}{c}{IR} & \multicolumn{3}{c}{Zero-shot} \\
        & R@1 & R@5 & R@10 & R@1 & R@5 & R@10 \\
        \midrule
        w/o ITM-hn & 60.2 & 86.2 & 92.3 & 43.3 & 74.0 & 82.9  \\
        InterBERT & 61.9 & 87.1 & 92.7 & 49.2 & 77.6 & 86.0 \\
        % Two-stage & 70.3  \\
        % Out+In & MSM+MRM & - \\
        % Out+In+VCR & MSM+MRM & - \\
    \bottomrule
  \end{tabular}
\end{table}

\vpara{The Effects of ITM with Hard Negatives}
We also evaluate the effect of our designed image-text matching. Specifically, it strongly impacts the model performance on image retrieval and its zero-shot version. Thus we conduct the ablation study on the two tasks to figure out whether it can bring significant promotion. 
Table~\ref{tab:ablation-itm} demonstrates the results of the evaluation. It can be found that our designed ITM with negative sampling based on TF-IDF is able to significantly enhance the performance on the retrieval tasks. The model trained with our ITM can outperform the baseline on caption-based image retrieval by $+1.7$ in R\@1, $+0.9$ in R\@5, and $+0.4$ in R\@10, and it also outperforms the baseline on the zero-shot retrieval by $+5.9$ in R\@1, $+3.6$ in R\@5, and $+3.1$ in R\@10. 
These results demonstrate that the matching task with higher difficulty in differentiating positive and negative samples contributes to the learning through pretraining. Specifically, the boost in the zero-shot retrieval proves that higher difficulty for image-text matching effectively impacts cross-modal understanding. 

\begin{table*}[tb]
  \caption{Results on the GLUE dev set. We evaluate the performance of the BERT-base model, a single-stream model, and \interbert~on 8 tasks of GLUE. The results show that \interbert~can rival the BERT-base model in the tasks of natural language understanding. We report F1 scores for QQP and MRPC, Spearman correlations for STS-B, and accuracy scores for the rest.}
  \label{tab:glue}
  \centering
  \begin{tabular}{cccccccccc}
    \toprule
        Model & QNLI & CoLA & SST-2 & STS-B & RTE & MNLI (m/mm) & QQP & MRPC & Avg. \\
        \midrule
        BERT-base & \textbf{91.5} & 56.7 & \textbf{93.2} & 88.2 & \textbf{65.0} & 83.7 / \textbf{84.1} & 87.9 & \textbf{89.6} & \textbf{82.0} \\
        Single Stream & 90.8 & 52.3 & 91.6 & 88.6 & 59.2 & 82.6 / \textbf{84.1} & 87.8 & 86.4 & 80.0 \\
        \interbert  & 91.1 & \textbf{57.3} & 92.3 & \textbf{88.9} & 64.3 & \textbf{84.1} / 83.7 & \textbf{88.1} & 88.6 & 81.8 \\
        % Out+In & MSM+MRM & - \\
        % Out+In+VCR & MSM+MRM & - \\
    \bottomrule
  \end{tabular}
\end{table*}

\vpara{Performances in the Single-Modal Tasks} While multi-modal pretraining demonstrates effects in the aforementioned downstream tasks, it is still a question whether it still preserve the knowledge of single-modal representation and whether it can still achieve comparable performances in the single-modal tasks. To evaluate the model's robustness, we conduct an experiment on 8 tasks of GLUE~\citep{glue}, including QNLI, CoLA, SST-2, STS-B, RTE, MNLI, QQP, and MRPC.
% \footnote{We provide the details of the GLUE datasets in Appendix~\ref{sec:appendix_glue}.} 
We compare \interbert~with BERT-base and the single-stream multi-modal pretrained model (a simple BERT architecture).
% \footnote{This is a single-stream model, which is a BERT that processes the information of multiple modalities. We pretrain the model with the same data.} 
% \begin{table}[tb]
%   \caption{Results on the GLUE dev set. We evaluate the performance of the BERT-base model, a single-stream model, and \interbert~on 8 tasks of GLUE. The results show that \interbert~can rival the BERT-base model in the tasks of natural language understanding. We report F1 scores for QQP and MRPC, Spearman correlations for STS-B, and accuracy scores for the rest.}
%   \label{tab:glue}
%   \centering
%   \resizebox{\textwidth}{!}{
%   \begin{tabular}{cccccccccc}
%     \toprule
%         Model & MNLI (m/mm) & QQP & QNLI & MRPC & CoLA & SST-2 & STS-B & RTE & Avg. \\
%         \midrule
%         BERT-base & 83.7 / \textbf{84.1} & 87.9 & \textbf{91.5} & \textbf{89.6} & 56.7 & \textbf{93.2} & 88.2 & \textbf{65.0} & \textbf{82.0} \\
%         Single Stream & 82.6 / \textbf{84.1} & 87.8 & 90.8 & 86.4 & 52.3 & 91.6 & 88.6 & 59.2 & 80.0 \\
%         \interbert & \textbf{84.1} / 83.7 & \textbf{88.1} & 91.1 & 88.6 & \textbf{57.3} & 92.3 & \textbf{88.9} & 64.3 & 81.8 \\
%         % Out+In & MSM+MRM & - \\
%         % Out+In+VCR & MSM+MRM & - \\
%     \bottomrule
%   \end{tabular}
%   }
% \end{table}
%调整了列顺序

\begin{figure}
    \centering
    \includegraphics[width=0.9\linewidth]{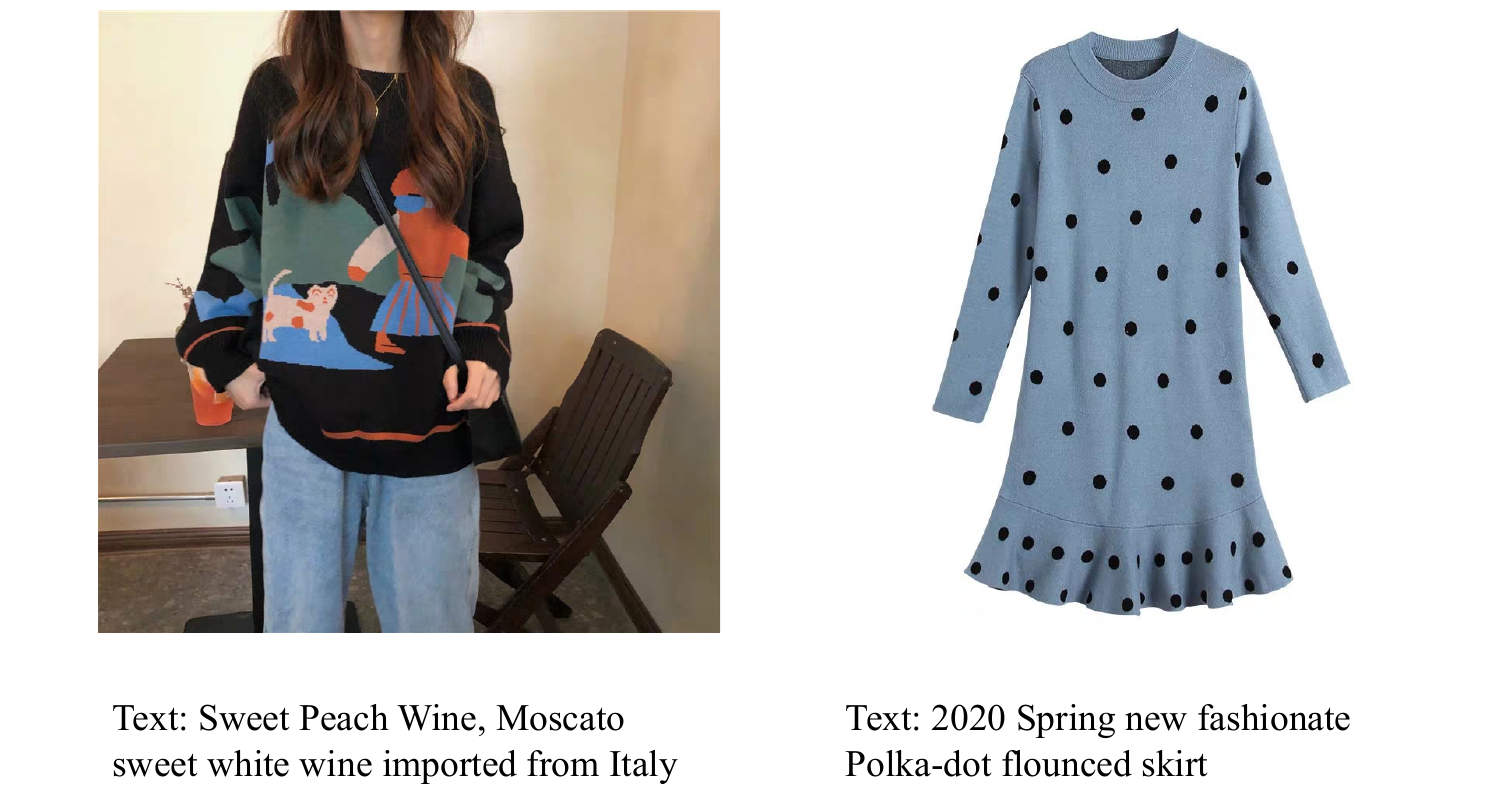}
    \caption{An illustration of two examples of the extracted image-text pairs from the mobile Taobao (the texts are translated from Chinese). Each image and its corresponding text describe a specific product. Both the images and texts are provided by the sellers themselves.}
    \label{fig:example_taobao}
\end{figure}

From Table~\ref{tab:glue}, it can be found that \interbert~can achieve similar performances compared with BERT-base (Avg: 82.0 vs 81.8), and it significantly outperforms the single-stream model without the two-stream extraction module (Avg: 81.8 vs 80.0). This indicates that \interbert~with the extraction module preserves the ability to model single-modal representations and it can adapt to single-modal downstream tasks without significant performance decrease. 

\vpara{The Effects of Initialization} In our experiments, we surprisingly find that the different weight initialization for pretraining has different impacts on the finetuning on different downstream tasks. As mentioned above, we initialize a part of our model with the weights of the pretrained BERT-base model. Here we compare the models with or without BERT initialization on the performances on the downstream tasks. While we find that such initialization has little impacts on the retrieval tasks, we also find that the initialization is surprisingly significant to the finetuning on VCR. The model without BERT initialization suffers from a severe performance downgrade. Its performances in Q$\rightarrow$A and QA$\rightarrow$R are 65.3 (-10.7\%) and 64.4 (-13.9\%), and VilBERT without BERT initialization performs worse (61.7 in Q$\rightarrow$A and 59.7 QA$\rightarrow$R). This demonstrates the importance of the pretrained NLP model on the multi-modal tasks concerned with reasoning as it can improve the effects of text processing and thus enhance its language understanding capability. It also shows that sufficient interaction between modals through all-attention can alleviate the problem. However, this can be a starting point for the research in the effect of initialization on multi-modal pretraining.

\section{Related work}
\label{sec:related-work}

In this section, we review the studies in pretraining methods, especially the pretraining in NLP and multi-modal pretraining. 

\vpara{Single-Modal Pretraining}
% Before the burst of pretraining in NLP, which has significant influence in the community of deep learning and artificial intelligence, the research in computer vision has long been dependent on the pretraining on large-scale datasets. The most famous one for pretraining is ImageNet~\cite{imagenet}. A series of model architectures for CV tasks, for example image classfication, are pretrained on the dataset. A typical one is AlexNet~\cite{alexnet}, which demonstrated the tremendous power of neural networks compared with conventional machine learning algorithms. Later, VGG~\cite{vgg} and ResNet~\cite{resnet} were also pretrained on the ImageNet and became the commonly used backbones for most downstream tasks of computer vision. Relatively, pretraining in NLP had long been lagging behind.
The recent years have witnessed the development of the pretraining in NLP. ELMo~\cite{elmo}, which is an LSTM-based~\cite{lstm} language model, has attracted the attention of NLP researchers as it demonstrated that pretraining is also available for NLP tasks. Later, ULMFit~\cite{ulmfit} proposed some techniques and gained improvements in several downstream tasks. Yet these models are based on the conventional recurrent neural network architecture. GPT~\cite{gpt} is the first language model based on Transformer~\cite{transformer} architecture, which is a unidirectional decoder. In concern of full observation of the context, \citet{bert} proposed BERT, a bidirectional encoder based on Transformer. BERT reached state-of-the-art performances in a number of NLP downstream tasks, including natural language inference~\cite{glue} and question answering~\cite{squad}. There have been a series of studies following BERT~\cite{roberta, xlnet, albert}. These models have achieved superior performances over the baselines and some even outperformed human performance. 

\vpara{Multi-modal Pretraining}
The success of pretraining in NLP raised the attention in multi-modal pretraining. VideoBERT~\cite{videobert} is regarded as the first work in multi-modal pretraining. It is a model pretrained on the extracted video frame features and texts. A contemporaneous work of VideoBERT is CBT~\cite{cbt}, which is also pretrained on video-text pairs. \citet{miech2020end} leveraged unlabeled narrated videos for video representation learning.

Inspired by the starting work in multi-modal pretraining, more researchers have turned their focus to visual-linguistic pretraining. 
There are mainly two streams of model architectures for this task. One is the single-stream model~\cite{b2t2, uniter, unicoder-vl, visualbert, vl-bert, gan2020large, oscar, vlp}. 
\citet{unicoder-vl} processed the concatenation of objects and words with a BERT model and pretrain it with the three conventional tasks. \citet{uniter} and \citet{qi2020imagebert} proposed similar methods but with more pretraining tasks and larger datasets. \citet{gan2020large} further improved the model with adversarial training strategy. \citet{vl-bert} used identical architecture but they pretrained the object detector and added single-modal data. \citet{huang2020pixel} attempted to input pixels directly instead of detected objects. \citet{oscar} leveraged the object labels to enhance the cross-modal alignments. \citet{vlp} proposed a unified single-stream model which jointly learns the caption generation and VQA tasks.

The other form of the model architecture is the two-stream model~\citep{lxmert, vilbert, vilbert-mt, yu2020ernie}. \citet{lxmert} proposed a two-stream model with co-attention and pretrained the model only with the in-domain data. \citet{vilbert} proposed a similar architecture with a more complex co-attention, and pretrained the model with the out-of-domain data, and \citet{vilbert-mt} further improved VilBERT with multi-task learning. Recently, \citet{yu2020ernie} incorporated the scene graph into the model, which brought performance gains. Aside from these works, \citet{singh2020we} discussed the impact of choosing the pretraining datasets on the performance of the downstream tasks.

In this work, we simply focus on the design of architecture and pretraining tasks. 
The single-stream models mostly apply BERT to multi-modal pretraining in a straightforward fashion, while the two-stream models have respective encoders for modalities and a co-attention module for the cross-modal interaction. 
These models either lack the independence of each modality or lack sufficient interaction across modalities. Furthermore, there is still room for setting training tasks for more effective pretraining. 
Compared with the previous work, our proposed method has several significant differences. Our proposed model architecture is effective in capturing modal interaction with an all-attention-based module and obtaining modal independence with the two-stream extraction module. Besides, our proposed masked group modeling improves the model's ability to predict a span or a region, so that the model can be more effective. 
% Furthermore, we first pretrain the model on basic ``out-of-domain'' datasets and further pretrain on the ``in-domain'' datasets. 

\section{Conclusion}
\label{sec:conclusion}

In this paper, we propose a new approach for multi-modal pretraining, \interbert. The model architecture consists of a single-stream interaction module for sufficient interaction and a two-stream extraction module for the separation of modal information. 
Furthermore, to strengthen its ability of modeling image and language, we pretrain the model with the tasks of MGM and ITM-hn. Experimental results demonstrate that our \interbert~can outperform the baselines and rival the recent multi-modal pretrained models in the downstream tasks, and our online deployment for recommendation shows its advantages in CTR and exposed category width over the single-modal baseline. 
The analyses show that the pretraining tasks can enhance the model performance, and \interbert~can adapt to single-modal tasks without significant performance downgrade. 
Also, we find out that the weight initialization for pretraining makes a difference to downstream tasks. 
We hope this study can provide some insights into multi-modal pretraining, and in the future, we will endeavor to figure out better model architecture and training tasks for the improvement in multi-modal representation learning. 

\bibliographystyle{ACM-Reference-Format}
\bibliography{sample-base}

%%
%% If your work has an appendix, this is the place to put it.
\appendix

\section{Appendix}
\label{sec:appendix}

\subsection{Data statistics}
\label{sec:appendix_data}
% \vpara{Pretraining Datasets of Image-caption Pairs} 
\vpara{Pretraining Datasets}
The image caption datasets for pretraining are Conceptual Caption (CC)~\cite{cc}, SBU Captions~\cite{sbu}, and COCO Captions~\cite{coco}. The detailed data statistics are demonstrated in Table~\ref{tab:pretrain_dataset}. In CC and SBU, each image is paired with a text as its description, while in COCO, there are around 5 texts that describe the same text. We also provide the number of images in COCO in the table. 

% \vpara{Pretraining Dataset of Product Image-title Pairs} 
% Since the images of the KDD Cup retrieval task are in very different domain compared with other downstream tasks considered in this work, we specially construct a pretraining dataset for this task. 
% Over 200 million product images are crawled from \textit{taobao.com}, which are homogeneous to the images in the KDD Cup task. Though these products are not paired with queries like the downstream samples, we can still use their raw product tities for pretraining.

\begin{table}[tb]
  \caption{Data statistics of the datasets for pretraining. The numbers in the parentheses refer to the numbers of images.}
  \label{tab:pretrain_dataset}
  \centering
  \begin{tabular}{ccc}
    \toprule
        Datasets&Training&Validation\\
        \midrule
        Conceptual Caption & 3.3M & 14K\\
        SBU & 890K & 10K \\
        COCO & 587K (117K) & 15K (3K) \\
    \bottomrule
  \end{tabular}
\end{table}

\begin{table}[tb]
  \caption{Data statistics of the datasets of the downstream tasks. ``i'' refers to the number of images, and ``t'' refers to the number of texts.}
  \label{tab:downstream_datasets}
  \centering
  \begin{tabular}{cccc}
    \toprule
        Datasets & Training & Validation & Testing\\
        \midrule
        Flickr30K & i:29K, t:145K & i:1K, t:5K & i:1K, t:5K \\
        % VQA & i:83K, t:444K & i:41K, t:214K & i:81K, t:448K \\
        VCR & i:80K, t:213K & i:10K, t:27K & i:10K, t:25K \\
        % RefCOCO+ & i:20K, t:141K & - & - \\
        % KDD Cup & i:3M, t:3M & i:15K, t:0.5K & i:30K, t:1K \\
    \bottomrule
  \end{tabular}
\end{table}

\vpara{Downstream Datasets} 
We demonstrate the detail data statistics of the datasets for finetuning in Table~\ref{tab:downstream_datasets}. The numbers of image and text of each dataset are provided. 
% Flickr30K~\citep{flickr} is the dataset for caption-based image retrieval as well as its zero-shot version, and VCR is the one for visual commonsense reasoning. 

\vpara{Dataset for Online Deployment}
% \label{sec:appendix_taomultimodal}
While our deployment relies on the large-scale dataset, we select a part containing around 3m samples for release, which is called TaoMultimodal. 
Here we provide more details about the data construction and preprocessing. 

% We collect the data for multi-modal pretraining in Chinese from the mobile Taobao. In the mobile Taobao, each product has at least one main image (the first image of the product, which is specified by the seller) and one product title. The image demonstrates the main features of the product and the title describes its basic information. Both the image and text are provided by the seller. 
% We extract image-text pairs from the best-selling products under the categories ``women clothes'' of the mobile Taobao. For the data for finetuning, we extract the image-text pairs from the products under the categories ``women shoes''. 

% \vpara{Preprocessing}
We apply the same preprocessing methods to both datasets. The preprocessing includes object detection for object representations and data cleaning for texts. We obtain object representations by using an object detector based on Faster-RCNN, which is trained on the data of the mobile Taobao. There are $33$ categories for object classification. We extract the bounding boxes with confidence scores larger than $0.1$, and we obtain no more than $16$ objects for each image.\footnote{Unlike the other datasets, the images in TaoMultimodal contain relatively fewer objects.} The size of the object representations is $2048$. As to the data cleaning for text, we first remove the titles without any Chinese character. Moreover, we conduct word segmentation with the AliNLP tool\footnote{https://data.aliyun.com/product/nlp} on the texts, and truncate the text by words in order to make sure the text is no longer than $36$ characters. Furthermore, we remove those titles that trigger our spam detector, including texts that are concerned with pornography, abuse, politics, terrorism, etc. 
Finally, we obtain a dataset of 3.1M image-text pairs for pretraining and 200K for finetuning.

\subsection{Implementation details}
\label{sec:appendix_impl}

In the following, we introduce the details of our implementation in pretraining and finetuning on each downstream task, including the model architecture, optimizer, hyperparameters, etc. 

\vpara{Pretraining}
Here we provide the experimental details about our implementation for pretraining. 
The object representations of the images as well as their bounding boxes are generated by an object detector based on Faster R-CNN~\cite{faster-rcnn} with a backbone of ResNet-101~\cite{resnet}, which is trained on Visual Genome~\cite{visual_genome}. This detector is applied for the bottom-up top-down attention model for image captioning~\cite{bottom-up}, and we downloaded the pretrained detector from their provided link.\footnote{https://github.com/peteanderson80/bottom-up-attention} For the text processing, we tokenize the texts with BERT's tokenizer and directly use BERT-base's embedding layer for word embedding. The vocabulary size is $30522$ and the embedding size is $768$. For the consistency between word embedding and object representation, we transform the object representations of $2048$ dimensions to $768$ through MLP. 

The hidden size of multi-head attention is also set to $768$. The number of attention head is $12$. For the FFN, both the input and output sizes are $768$ for stacking layers, and the intermediate size is $3072$. As to the LN layer inside each layer, we use BERT's LN with $e=1e-12$. The single-stream interaction module consists of $12$ layers of Transformer layer, and its weight parameters are initialized with the pretrained BERT-base model. The two-stream independence module contains two Transformers for both modalities on top of the single-stream interaction module. Each has $6$ layers of Transformer layer. The new weight parameters are randomly initialized based on the Gaussian distribution of zero mean and standard deviation of $0.02$, following~\citet{bert}. 
We pretrain the model with AdamW~\citep{adamw} whose initial learning rate of $1e-4$, $\beta_1=0.9$, $\beta_2=0.9999$, $e=1 \times 10 ^ {-6}$ and a weight decay of $0.01$. We apply the linear decay learning rate scheduler with a warm-up period of $10000$ steps. 
The batch size for training is $512$. 
For the pretraining dataset of image-caption pairs, we pretrain our \interbert~on 8 V100 GPUs for $20$ epochs. 

\vpara{Finetuning}
For the finetuning on Flickr30K image retrieval, the maximum number of objects is $100$ and the actual numbers are between $90$ and $100$. 
% We use an initial learning rate of $4 \times 10 ^ {-5}$ and a batch size of $32$. 
The model reuses the output layer of the pretraining ITM task to compute the matching scores. We finetune the model with a batch size of $32$ and train it on 8 V100 for $20$ epochs. We use AdamW optimizer with an initial learning rate of $4 \times 10 ^ {-5}$ and apply a linear decay learning rate scheduler with a warmup period of $10000$ steps. We finetune the model for $20$ epochs. 
For the finetuning on VCR, we use similar hyperparameters with those in the finetuning on Flickr30K, but we use a smaller learning rate $2 \times 10 ^ {-5}$ and a smaller batch size $32$, and we only finetune the model for $5$ epochs. 
% For the finetuning on the KDD Cup retrieval task, since there are relatively fewer objects in commodity images, we set the maximum number of objects to $36$ to reduce the memory cost. The structural setting of the output layer is similar to the Flickr30K task. We use an initial learning rate of $5 \times 10 ^ {-6}$ and batch size $288$. The model is also finetuned for $20$ epochs. 
Furthermore, we apply exponential moving average with a rate of $0.9999$ on the finetuned models for the final model, so that it can be more robust and reach better performance in testing.

\vpara{Deployment}
We store the representation vector from the multi-modal pretrained model and the BERT-based baseline for each item of the item pool. We then send them into a vector-based KNN service for nearest neighborhood search. For each trigger item, we collect the top $5$ nearest neighbor items offline. This decision of hyperparameter is based on our preliminary online experiments. The online service receives the triggers from the user histories and search the recalled candidate items accordingly. This query service requires around 10ms. A separate ranking system is responsible for the production of final results. Both of our methods share the same preprocssing the postprocessing. For the A/B testing, the A/B buckets adds a trail of multimodal recall and that of single-modal recall respectively. This is the only difference between the buckets. Each of them shares at least 5\% traffic and lasts at least 7 days. 

\vpara{Hardware Configuration}
The experiments are conducted on a Linux server equipped with an Intel(R) Xeon(R) Platinum 8163 CPU @ 2.50GHz, 512GB RAM and 8 NVIDIA V100-SXM2-16GB GPUs.
% Our full set of experiments took about 3 days with this multi-GPU setting.

\vpara{Software}
The experiments are implemented in python 3.6 and PyTorch 1.1.0~\cite{paszke2017automatic}. The code is based on Transformers~\cite{transformers}.\footnote{https://github.com/huggingface/transformers}
% We release our code for preprocessing the dataset and running all the experiments.

\subsection{Details of the GLUE tasks}
\label{sec:appendix_glue}

The GLUE benchmark~\cite{glue} consists of a series of NLP tasks, including QNLI, CoLA, SST-2, STS-B, RTE, MNLI, QQP, and MRPC,. We use them to evaluate the robustness of \interbert~in single-modal downstream tasks.

\vpara{QNLI} Question Natural Language Inference is a binary classification task of SQuAD (Stanford Question Answering Dataset) \citep{squad, glue}. It requires the model to judge whether the given answer is a correct one of the given question in a sentence pair. 

\vpara{CoLA}  The Corpus of Linguistic Acceptability~\cite{cola} is a task of binary sentence classification. It requires the algorithms to check whether an English sentence is linguistically acceptable (grammatical and consistent with the world knowledge). 

\vpara{SST-2} The Stanford Sentiment Treebank~\cite{sst-2} is a task of binary sentiment classification. It requires the algorithms to check whether a sentence is positive or negative. The sentences are extracted from movie reviews with human annotations.

\vpara{STS-B} The Semantic Textual Similarity Benchmark~\cite{sts-b} is a task of classification of semantic similarity. The sentences are extracted from news headlines and other sources. The algorithms should learn to score two sentences from $1$ to $5$ for their semantic similarity. 

\vpara{RTE} Recognizing Textual Entailment~\cite{rte} is a task of natural language inference. This task provides a sentence pair and requires the algorithms to check the relations between the sentences, including ``entailment'', ``contraction'' and ``neutral''.

\vpara{MNLI} Multi-Genre Natural Language Inference~\citep{mnli} is a task of entailment with a large dataset. It requires the algorithm to figure out the relation of a pair of sentences. The relations include ``entailment'', ``contradiction'', and ``neutral''.

\vpara{QQP} Quora Question Pairs \citep{qqp} is a task to check if two questions are semantically identical. The questions are extracted from Quora\footnote{\url{http://quora.com/}}.

\vpara{MRPC} Microsoft Research Paraphrase Corpus is a dataset of sentence pairs from news websites. The task is to check whether two sentences are semantically identical. 

We truncate the input texts to ensure that the maximum length is $128$. The input texts are all lower-cased. We use a batch size of $128$ and a learning rate of $2 \times 10 ^ {-5}$. We finetune the model on 8 Nvidia V100 GPUs with gradient accumulation for $3$ epochs.

\end{document}